\documentclass[12pt,letterpaper]{article}
\usepackage{authblk} 
\usepackage[comma,authoryear]{natbib} 
\usepackage{amssymb}
\usepackage{latexsym}
\usepackage{graphicx}
\usepackage{url}
\usepackage{array}
\usepackage{color,soul} 
\usepackage{float} 
\usepackage[hidelinks,backref=page,bookmarksnumbered,pdfhighlight=/P]{hyperref} 
\usepackage[colorinlistoftodos]{todonotes}
\usepackage{fancyhdr} 
\usepackage{geometry}
\usepackage{layout}
\begin{document} 
\clearpage
\newpage 
\title{A New 2.5D Representation for Lymph Node Detection using Random Sets of Deep Convolutional Neural Network Observations}
\author[1]{\small Holger R. Roth\thanks{holger.roth@nih.gov, h.roth@ucl.ac.uk}}
\author[1]{\small Le Lu}
\author[1]{\small Ari Seff}
\author[1]{\small Kevin M. Cherry}
\author[1]{\small Joanne Hoffman}
\author[1]{\small Shijun Wang}
\author[1]{\small Jiamin Liu}
\author[1]{\small Evrim Turkbey}
\author[1]{\small Ronald M. Summers}
\affil[1]{\small Imaging Biomarkers and Computer-Aided Diagnosis Laboratory,\\
Radiology and Imaging Sciences, National Institutes of Health Clinical Center, Bethesda, MD 20892-1182, USA.}
\date{\small \today}
\maketitle
\begin{abstract} 
\noindent Automated Lymph Node (LN) detection is an important clinical diagnostic task but very challenging due to the low contrast of surrounding structures in Computed Tomography (CT) and to their varying sizes, poses, shapes and sparsely distributed locations. State-of-the-art studies show the performance range of 52.9\% sensitivity at 3.1 false-positives per volume (FP/vol.), or 60.9\% at 6.1 FP/vol. for mediastinal LN, by one-shot boosting on 3D HAAR features. In this paper, we first operate a preliminary candidate generation stage, towards $\sim$100\% sensitivity at the cost of high FP levels ($\sim$40 per patient), to harvest volumes of interest (VOI). Our 2.5D approach consequently decomposes any 3D VOI by resampling 2D reformatted orthogonal views $N$ times, via scale, random translations, and rotations with respect to the VOI centroid coordinates. These random views are then used to train a deep Convolutional Neural Network (CNN) classifier. In testing, the CNN is employed to assign LN probabilities for all $N$ random views that can be simply averaged (as a set) to compute the final classification probability per VOI. We validate the approach on two datasets: 90 CT volumes with 388 mediastinal LNs and 86 patients with 595 abdominal LNs. We achieve sensitivities of 70\%/83\% at 3 FP/vol. and 84\%/90\% at 6 FP/vol. in mediastinum and abdomen respectively, which drastically improves over the previous state-of-the-art work.
\end{abstract}
\section{Introduction}
Accurate detection and segmentation of enlarged Lymph Nodes (LNs) plays an important role for the staging of many diseases and their treatment, e.g. lung cancer, lymphoma and inflammation. These pathologies can cause affected LNs to become enlarged, i.e. swell in size. A LN's size is typically measured on Computed Tomography (CT) images following the RECIST guideline \citep{therasse2000new}. A LN is considered enlarged if its smallest diameter (along its short axis) measures more than 10 mm on an axial CT slice (see Fig. \ref{fig:lymph_node_image_patch}). Quantitative analysis plays a pivotal role for assessing the progression of certain diseases, accurate staging, prognosis, choice of therapy, and follow-up examinations. Radiologists need to detect, quantitatively evaluate and classify LNs. This assessment is typically done manually and is error prone due to the fact that LNs can vary markedly in shape and size and can have attenuation coefficients similar to those of surrounding organs (see Fig. \ref{fig:lymph_node_image_patch}). Furthermore, manual processing is time-consuming and tedious and might delay the clinical workflow. 

Previous work on computer-aided detection (CADe) systems for LNs mostly uses direct 3D information from volumetric CT images. State-of-the-art methods \citep{barbu2012automatic,feulner2013lymph} perform boosting-based feature selection and integration over a pool of $\sim$50 thousand 3D Haar-like features to obtain a strong binary classifier for detecting LNs. Due to the limited availability of annotated training data and the intrinsic high dimensionality, modeling complex 3D image structures for LN detection is non-trivial. Particularly, lymph nodes have large within-class appearance, location or pose variations, and low contrast from surrounding anatomies over a patient population. This results in many false-positives (FP), to assure a moderately high detection sensitivity \citep{feuerstein2009automatic}, or only limited sensitivity levels \citep{barbu2012automatic,feulner2013lymph}. The good sensitivities achieved at low FP range in \citet{barbu2012automatic} are not directly comparable with the other studies since \citet{barbu2012automatic} reports on axillary, pelvic, and only some parts of the abdominal regions, while others evaluate only on mediastinum \citep{feuerstein2012mediastinal,feulner2013lymph,feuerstein2009automatic} or abdomen \citep{nakamura2013automatic}. High numbers of FPs per image make efficient integration of CADe into clinical workflow challenging. 

Our method employs a LN CADe systems \citep{liu2014mediastinal,cherry2014abdominal} with high sensitivities as the first stage and focuses on effectively reducing FPs. In comparison, the direct one-shot 3D detection \citep{barbu2012automatic,feulner2013lymph} saturates at $\sim$65\% sensitivity at full FP range. Recently, the availability of large-scale annotated training sets and the accessibility of affordable parallel computing resources via GPUs has made it feasible to train deep Convolution Neural Networks (CNNs) and achieve great advances in challenging ImageNet recognition tasks \citep{krizhevsky2012imagenet,zeiler2013visualizing}. Studies that apply deep learning and CNNs to medical imaging applications also show promise, e.g. \citep{prasoon2013deep}, and classifying digital pathology \citep{cirecsan2013mitosis}. Extensions of CNNs to 3D have been proposed, but computational cost and memory consumption are still too high to be efficiently implemented on today's computer graphics hardware units \citep{prasoon2013deep,turaga2010convolutional}. In this work, we investigate the feasibility of using CNNs as a highly effective of FP reduction. We propose to use 3D VOIs with a new 2.5D representation that may easily facilitate a generally-purposed 3D object detection by classification scheme.
\section{Methods}
\subsection{LN Candidate Detection in Mediastinum and Abdomen}
\label{sec:cade}
We use a preliminary CADe system for detecting LN candidates from mediastinal \citep{liu2014mediastinal} and abdominal \citep{cherry2014abdominal} CT volumes. In the mediastinum, lungs are segmented automatically and shape features are computed at voxel-level. The system uses a spatial prior of anatomical structures (such as esophagus, aortic arch, and/or heart) via multi-atlas label fusion before detecting LN candidates using a Support Vector Machine (SVM) for classification. In the abdomen, a random forest classifier is used to create voxel-level LN predictions. Both systems permit the combination of multiple statistical image descriptors and appropriate feature selection in order to improve LN detection beyond traditional enhancement filters. LN candidate generation is not a core topic of this paper. Currently, 94\%-97\% sensitivity level at the rates of 25-35 FP/vol. can been achieved \citep{liu2014mediastinal,cherry2014abdominal}. Given sufficient training for the LN candidate generation step, close to 100\% sensitivities could be reached in the future.
\subsection{CNN training on 2.5D Image Patches}
\begin{figure}
	\centering
		\includegraphics[width=1.0\textwidth]{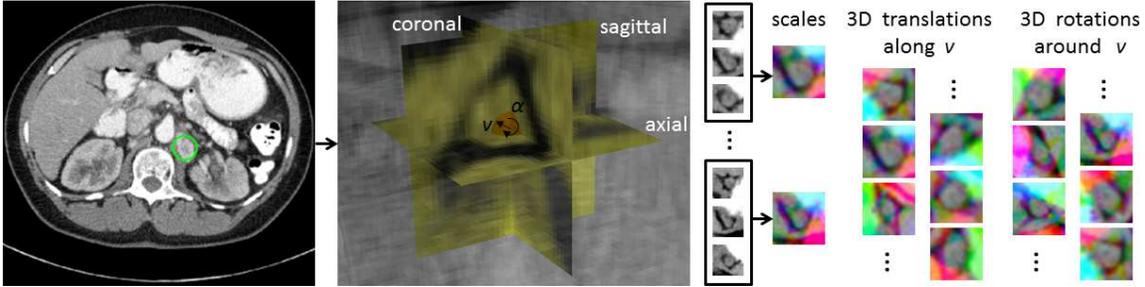}
	\caption{Examples of lymph nodes (circled) in an axial CT slice of the abdomen. Image patches are generated from CADe candidates, using different scales, 3D translations (along a random vector $v$) and rotations (around a random vector $v$ by a random angle $\alpha$). Each image patch (visualized as RGB image) is composed of an axial (R), coronal (G), and sagittal slice (B) and centered at a LN candidate.}
	\label{fig:lymph_node_image_patch}
\end{figure}

\begin{figure}
	\centering
		\includegraphics[width=1.0\textwidth]{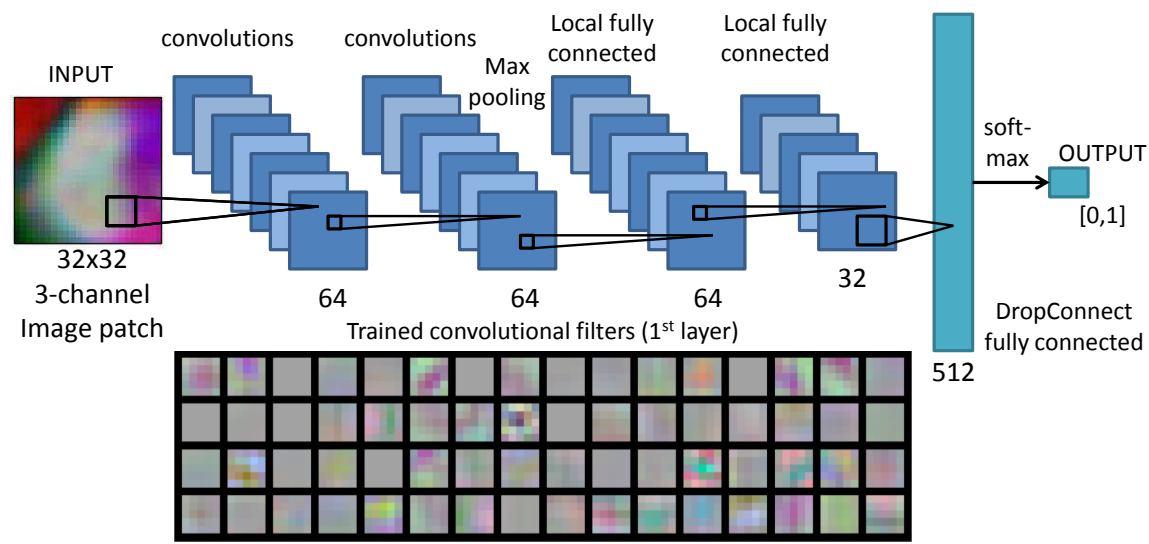}
	\caption{Our convolution neural network consists of two convolutional layers, max-pooling layers, locally fully-connected layers, a DropConnect layer, and a final 2-way softmax layer for classification. The number of filters, connections for each layer, and the first layer of learned convolutional kernels are shown.}
	\label{fig:CNN_layout}
\end{figure}

In general computer vision, a CNN is typically designed to classify color images that contain three image channels: Red, Green and Blue (RGB). We map this set-up by assigning the axial, coronal and sagittal slices in a Volume-of-Interest (VOI) into to these three channels (see Fig. \ref{fig:lymph_node_image_patch}). Our approach is similar to \citet{prasoon2013deep} in that we use the three orthogonal slices (axial, coronal and sagittal) through the center of a CADe mark as the input patch. However, we aim to simplify the training of the CNN by jointly using three channel images. This differs from the approach of \citet{prasoon2013deep} that uses three individual and separately trained CNNs on each one of the orthogonal image slices, with a subsequent fusion of their predictions for image segmentation. The 3D CT data is resampled in order to extract VOIs at $N_s$ different physical scales $s$ (the edge length of each VOI), but with fixed numbers of voxels. In order to increase the training data variation and to avoid overfitting (analogous to the 2D data augmentation approach in \citet{krizhevsky2012imagenet}), each VOI is also translated along a random vector $v$ in 3D space $N_t$ times. Furthermore, each translated VOI is rotated around a randomly oriented vector $v$ at its center $N_r$ times by a random angle $\alpha=\left[0^{\circ},\ldots,360^{\circ}\right]$, resulting in $N=N_s\times N_t\times N_r$ random observation of each VOI (similar to \citet{gokturk01astatistical}). This permits easy expansion of both the training and testing data for this type of neural net application. When classifying unseen data, the $N$ random CNN predictions can be simply averaged at each VOI to compute a per-candidate probability:
\begin{equation}
	p\left(x|\{P_1(x),...,P_N(x)\}\right) = \frac{1}{N}\sum_{i=1}^{N}P(x),
	\label{equ:prob}
\end{equation}
where $P_i(x)$ is the CNN's classification probability for one individual image patch. 
The main purpose of this approach is to decompose the volumetric information from each VOI into a set of random 2D images (with three channels) that combine orthogonal slices at $N$ reformatted orientations in 3D. Our relatively simple re-sampling of the 3D data circumvents using 3D CNN directly \citep{turaga2010convolutional}. This not only greatly reduces the computational burden for training and testing, but more importantly, alleviates the curse-of-dimensionality problem. Direct training 3D deep CNN \citep{turaga2010convolutional} for the volumetric object detection problem may not be feasible due to severe lack of sufficient training samples, especially in the medical imaging domain. CNNs generally need tremendous amounts of training examples to address overfitting, with respect to the large number of  parameters. \citet{krizhevsky2012imagenet} uses translational shifting and mirroring of 2D image patches for this purpose. Random resampling is an effective and efficient way to increase the amount of available training data. Our 2.5D representation is intuitive and applies the success of large scale 2D image classification, using CNN \citep{krizhevsky2012imagenet} effortlessly into 3D space. The above averaging process (i.e., Eq. \ref{equ:prob}) further improves the robustness and stability of 2D CNN labeling on random views (see Sec. \ref{sec:results}).
\paragraph{The CNN architecture} typically consists of several layers that apply convolutional filters to the input images (hence the name). The subsequent layers consist of max-pooling layers, fully-connected layers, and a final 2-way softmax layer for classification (see Fig. \ref{fig:CNN_layout}). In order to avoid overfitting, we use a recently published method called ``DropConnect'' that behaves as a regularizer when training the CNN \citep{wan2013regularization}. DropConnect is a variation of the earlier proposed ``DropOut'' method. In order to allow efficient training of the CNN, we use a GPU-based open-source implementation by \citet{krizhevsky2012imagenet} with the DropConnect extension by \citet{wan2013regularization}. Alongside the use of GPU acceleration, a speed-up in training has been achieved by using rectified linear units as the neuron model instead of the standard neuron model $f(x)=\tanh(x)$ or $f(x)=(1 + e^{-x})^{-1}$ \citep{krizhevsky2012imagenet}. At this time, the optimal architecture of CNNs for a particular image classification task is difficult to determine \citep{zeiler2013visualizing}. We evaluate several CNNs with slightly different layer architectures to choose the best CNN architecture for our classification task and find relatively stable behavior on our datasets. Hence, we fix the CNN architecture for the subsequent cross-validation performed in this study. A recent approach proposes to visualize the trained CNN model by deconvolution and in order aid understanding the behavior of CNNs \citep{zeiler2013visualizing}. These methods have the potential to allow better CNN design rather than using a heuristic approach as in this work.
\section{Evaluation and Results}
\label{sec:results}
\begin{figure}
	\centering
		\includegraphics[width=1.0\textwidth]{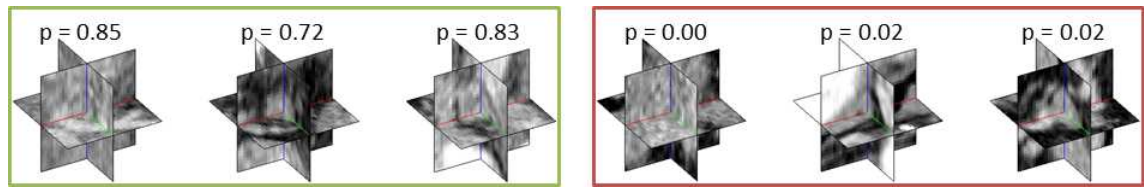}
	\caption{Test probabilities of the CNN for being a lymph node on `true' (left box) and `false' (right box) lymph node candidate examples.}
	\label{fig:lymphnode_CNN_predictions}
\end{figure}

\begin{figure}
	\centering
		\includegraphics[width=1.0\textwidth]{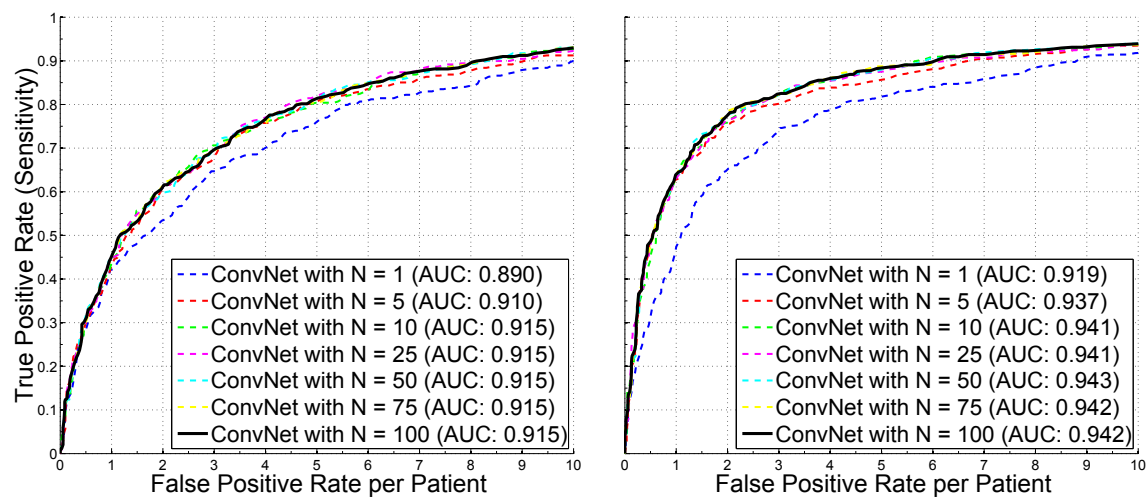}
	\caption{Free-response receiver operating characteristic (FROC) curves for a 3-folded cross-validation using a varying number of $N$ random CNN observers in 90 patients in the mediastinum (left) and 86 patients in the abdomen (right). AUC values are computed for corresponding ROC curves.}
	\label{fig:roc_froc}
\end{figure}

Radiologists labeled a total of 388 mediastinal LNs as ‘positives' in CT images of 90 patients and a total of 595 abdominal LNs in 86 patients. In order to objectively evaluate the performance of our CNN based 2.5D detection module, 100\% sensitivity at the LN candidate generation stage is assumed by injecting the labeled LNs into the set of CADe LN candidates (see Sec. 2.1). The CADe systems produce a total of 3208 false-positive detections ($>$15 mm away from true LN) in the mediastinum and 3484 in the abdomen. These false-positive detections are used as ‘negative' LN candidate examples for training the CNNs. All patients are randomly split into three subsets (at the patient level) to allow a 3-folded cross-validation. We use different sample rates of positive and negative image patches to generate a balanced training set. This proves beneficial for training the CNN -- no balancing is done during cross-validation. Each three-channel image patch is centered at a CADe coordinate with $32 \times 32$ pixels. All patches are sampled at 4 scales: $s=\left[30,35,40,45\right]$ mm for the VOI edge length in physical image space, after iso-metric resampling of the CT image (see Fig. \ref{fig:lymph_node_image_patch}). We use a soft-tissue window level of [-100, 200 HU] as in \citet{barbu2012automatic}. Furthermore, all VOIs are $N=100$ times randomly translated (up to 3 mm) and rotated at each scale ($N_s=4$, $N_t=5$ and $N_r=5$). We train separate CNN models for mediastinum and for abdomen. Training each CNN model takes 9-12 hours on a NVIDIA GeForce GTX TITAN, while running the 2.5D image patch classification for testing runs in only circa 5 minutes. Image patch extraction from one CT volume takes around 2 minutes. We then apply the trained CNN to classify image patches from the testing datasets. Figure \ref{fig:lymphnode_CNN_predictions} shows a typical classification probability on a random subset of test VOIs.
Averaging the $N$ predictions at each LN candidate allows us to compute a per-candidate probability $p(x)$, as in Eq. \ref{equ:prob}. Varying a threshold parameter on this probability allows us to compute the free-response receiver operating characteristic (FROC) curves. FROC curves are compared in Fig. \ref{fig:roc_froc} for varying amounts of $N$. It can be seen that the classification performance saturates quickly with increasing $N$. The classification sensitivity improves on the existing LN CADe systems \citep{liu2014mediastinal,cherry2014abdominal} from 55\% to 70\% in the mediastinum and from 30\% to 83\% in the abdomen at a low rate of 3 FP per patient volume (FP/vol.) at $N=100$. The area under the curve (AUC) improves from 0.76 to 0.942 in the abdomen, using the proposed false-positive reduction approach (AUC in the mediastinal was not available for comparison). At an operating point of 3 FP/vol., we achieve significant improvement: $p=7.6\times10^{-3}$ and $p=2.5\times10^{-14}$ in mediastinum and abdomen, respectively (Fisher's exact test). Further experiments show that performing a joint CNN model trained on both mediastinal and abdominal LN candidates together can improve the classification by $\sim$10\% to $\sim$80\% sensitivity improvement at 3 FP/vol. in the mediastinal set. The sensitivity level in the abdomen datasets remained stable. 
\section{Discussion and Conclusions}
This work (among others) demonstrates that deep CNNs can be generalized to 3D/2D medical image analysis tasks, such as effective FP reduction in CADe systems. Building upon existing methods for CADe of lymph nodes (LNs), we show that a random set of CNN observers (a 2.5D approach) can be used to reduce FPs, from the initial CADe detections. Different scales, sampling through random translations and rotations around each of the CADe detections can be exploited to prevent or alleviate overfitting during training and increase the CNN's classification performance. AUC and FROC exhibit significant improvement on sensitivity levels at the range of clinically relevant FP/vol. rates. These results are a drastic improvement compared to the state-of-the-art methods. \citet{feulner2013lymph} reports 52.9\% sensitivity at 3.1 FP/vol. in the mediastinum, while we achieve 70\% at 3 FP/vol. In the abdomen, the most recent work \citep{nakamura2013automatic} shows 70.5\% sensitivity at 13.0 FP/vol. We obtain 83\% at 3 FP/vol. (assuming  $\sim$100\% sensitivity at the LN candidate generation stage). Note that any direct comparison to another recent work is difficult since there are no common datasets available at the moment. Therefore, we will make our data\footnote{\url{http://www.cc.nih.gov/about/SeniorStaff/ronald_summers.html}} and supporting material\footnote{\url{https://sites.google.com/site/holgerrroth}} publicly available for convenient future comparison. 

The performance improvement using joint training on mediastinum and abdominal lymph nodes shows that it is beneficial for CNN to have larger, more varied and comprehensive datasets (which is coherent to the computer vision literature \citep{krizhevsky2012imagenet}). A companion approach \citep{seff2014viewaggregation} exploits an alternative shallow hierarchy for LN classification, using a view-level classification score aggregation by another classifier. While they show that this is helpful to achieve better FROC curves in their scheme, we find that the same sparsely weighted fusion via learning does not improve over the simple average of Eq. \ref{equ:prob}. This probably indicates the high quality of our deep CNN predictions and shows this approach to be very effective and efficient. Future work will investigate more sophisticated methods of label fusion from the CNNs. The proposed 2.5D generalization of CNNs shows promise for a variety of applications in computer-aided detection of 3D medical images. For future work, the 2D views with the highest probability of being a LN could be used to present reformatted visualizations at that orientation (optimal to the CNN) to assist in radiologists' reading. 
\paragraph{\textbf{Acknowledgments}}
This work was supported by the Intramural Research Program of the NIH Clinical Center. The final publication will be available at Springer.
\bibliographystyle{chicago}
\bibliography{HRRoth2014_lymphnode_cnn}
\end{document}